\documentclass[11pt]{article}

\usepackage[margin=1.1in]{geometry}
\usepackage{times}
\usepackage[T1]{fontenc}
\usepackage[utf8]{inputenc}
\usepackage{microtype}
\usepackage{amsmath}
\usepackage{amssymb}
\usepackage{booktabs}
\usepackage[round]{natbib}
\usepackage{url}

\newcommand{\bl}{b_L}
\newcommand{\mhat}{\hat{m}}

\title{Transformer MLP Gate Thresholds Are Couplings
to a Carried Reference Direction}
\author{Olli Tuomi\\Evident Solutions Oy}
\date{September 2026}

\begin{document}
\maketitle

\begin{abstract}
The corpus-mean direction of a transformer's residual stream is a
component shared across all inputs, and is commonly removed by
mean-centering before representational analysis. We present evidence
that it is a functional component: the reference against which the MLP
gate population sets its operating point. In Phi-2, an exact
decomposition of resting gate pre-activations shows that at mid-stack
layers the resting inhibition of $>$99.9\% of gates is carried by the
coupling $w\cdot\bl$ to the carried mean direction, at $48$--$56\times$
the explicit bias parameter, and the coupling is direction-specific: a
random direction at matched norm orders the population's firing rates at
Spearman $\rho \leq 0.09$ (median over 50 draws; up to $0.22$ in the tail) where
the reference reaches $0.95$. (That $0.95$ is near-tautological on its
own; \S\ref{sec:bias} derives its null.) Causally, removing the stream's
projection on the reference multiplies above-threshold firing by
$\sim$9$\times$, dose-monotonically, at $23$--$59\times$ a norm-matched
control; a random-initialised twin is flat, and replacement tests show
that the direction carries the function and the magnitude does not. A
direction-matched control makes the same point: noise
injected along the reference costs $8$--$42\times$ the same energy
along a random direction. The decomposition replicates on four further families
spanning both gate types (GELU with an explicit gate bias, bias-free
SwiGLU), and the dose-response on two of them. Across an eight-model
scan the mechanism is present in every GELU and SiLU family and absent
only in OPT, where an opposing LayerNorm bias cancels the carried
reference. Gate thresholds are implemented as couplings to a constant
the network builds for the distribution it is reading, with the bias
parameters contributing little; how much of that constant is carried in
the stream and how much in parameters depends on the architecture.
\end{abstract}

\section{Introduction}

Transformer residual streams contain large, stable components that are
shared across inputs: massive activations \citep{sun2024}, which give
rise to attention sinks \citep{xiao2023,sun2024}, and outlier dimensions
\citep{kovaleva2021,timkey2021}, whose magnitude tracks token frequency
\citep{puccetti2022}. Existing work
locates these objects and shows they matter for quantisation and
attention. What they are \emph{for}, computationally, is less settled.

This paper studies one such object, the corpus-mean direction of the
layer-normalised residual stream, $\bl$, and measures a specific
function for it: it is the reference against which the MLP gate
population sets its operating point. The claim decomposes into parts
that were tested separately, each against a random-initialised control
model (``twin'') of identical architecture:

\begin{enumerate}
\item \textbf{Object} (\S\ref{sec:object}): $\bl$ is constant across
  text for a given input distribution, is held at a fixed fraction of
  the growing stream, and, in Phi-2, rotates slowly with depth as a band of
  related directions.
\item \textbf{Function} (\S\ref{sec:function}): removing or amplifying
  the stream's projection on $\bl$ releases or suppresses gate firing,
  dose-monotonically, far beyond norm-matched controls; the twin is
  flat.
\item \textbf{Identity} (\S\ref{sec:identity}): the function is
  carried by the direction, not by magnitude.
\item \textbf{Implementation} (\S\ref{sec:bias}): resting gate
  inhibition is the coupling $w\cdot\bl$, not the bias parameter. This
  is an exact decomposition, not a fit.
\item \textbf{Persistence and dependence} (\S\ref{sec:maintenance}):
  the direction is continuously re-supplied by aligned per-layer mean
  writes, so it re-accumulates after local removal; defeating that
  restoration by co-removal is near-catastrophic for prediction.
\item \textbf{Development} (\S\ref{sec:formation}): the operating point
  is built early: on Pythia-410M it forms in the induction-transition
  window, a finer 160M grid shows it forms first, and it forms even with
  attention disabled (one 160M seed), so the token-local prediction task
  alone is enough to build it.
\end{enumerate}

These experiments are run on Phi-2 \citep{abdin2023phi}, with
Development on Pythia-410M \citep{biderman2023} checkpoints and two small
trained models. The two most portable tests, the static decomposition
(\S\ref{sec:bias}) and the causal dose-response (\S\ref{sec:function}),
are then tested on further families; the remaining causal tests are
Phi-2-only (Limitations).

Throughout, ``gate'' means the scalar nonlinearity of an MLP hidden
unit; ``duty cycle'' is the fraction of positions at which a unit's
activation is positive. Two facts motivate the hypothesis.
First, Phi-2's median duty cycle over 400 sampled gates is 0.076, against
0.48 at random init: trained gates are off by default, in line with the
known activation sparsity of trained transformer MLPs
\citep{li2023lazy,szatkowski2025},
and ``off by default'' is only defined relative to what occupies the
stream at rest. Second, take a gate's pre-activation at the positions where it fires
most strongly and ask what share of it comes from the stream's
corpus-mean component (the \emph{reference share of drive}; negative
values mean that component opposes firing). Its median over gates is
about $-0.40$ pooled over the eight layers measured ($-0.31$ to $-0.55$
per layer), and $\approx 0$ in the twin ($-0.11$ to $+0.01$): trained
gates are held down by something constant. The experiments below ask
whether that constant is functional structure.

\paragraph{Models.} Primary model: Phi-2 \citep{abdin2023phi} (2.7B, 32
layers, width $d{=}2560$), fp16, hooks on public weights; twin: same
architecture, random init. Cross-model replication:
GPT-2-medium \citep{radford2019gpt2}, Pythia-1.4B \citep{biderman2023},
Qwen2.5-1.5B \citep{qwen2024}, and TinyLlama-1.1B
\citep{zhang2024tinyllama} carry the static decomposition, and
Pythia-1.4B and Qwen2.5-1.5B also the causal dose-response; the static
decomposition further runs across an eight-model GELU/ReLU/SiLU scan
comprising those four plus Pythia-410M, SmolLM2 \citep{allal2025smollm},
and two OPT models (opt-350m, opt-1.3b) \citep{zhang2022opt}. Each is compared against its own
twin (same configuration, random weights). Development (\S\ref{sec:formation}):
Pythia-410M-deduped public checkpoints \citep{biderman2023}, plus a
160M model (Pythia-160M architecture, standard GELU MLP) trained here on
the Pile with dense early checkpoints, and an attention-disabled variant
of it (each layer's attention output projection zeroed and frozen).

\section{The object}
\label{sec:object}

\paragraph{Definitions.} $m_L$ is the corpus-mean raw residual at the
input of layer $L$; $\bl$ is the corpus mean of $x_{\text{ln}}$, the
layer-normalised MLP input at $L$ (the object gates actually read); $\mhat_L$ the unit
vector of $m_L$. The two are closely aligned ($\cos(\bl, m_L) =
0.81$--$0.83$ at L6--L18), so manipulations along $\mhat_L$ act on the
reference. Statistics are computed over document positions
$\geq 16$, excluding the attention-sink region.

\paragraph{Constancy.} Split-half cosine of $\bl$ across document
halves is $0.94$--$0.99$ at every measured layer. The constancy holds
for a given input distribution: the cosine between the reference
estimated on prose and on other inputs is $0.97$--$0.98$ for held-out
prose, $0.76$--$0.84$ for code, and $0.02$--$0.05$ for a repeated token,
where the twin keeps one direction for prose, code and random tokens
($\geq 0.99$) and only partly separates the repeated token
($0.51$--$0.52$). The mean also hides some
finer structure: token classes such as digits, sub-word continuations
and word-initial tokens have their own means, separable from $\bl$
above a split-half floor and partly as a trained property, but the class
partition accounts for only $2$--$4\%$ of the variance around the mean.

\paragraph{A constant fraction of a growing stream.} The
residual stream grows several-fold across the stack, so a fixed
coupling could only give a fixed operating point if something
normalised it. Over L2--L28, mean $\|x\|$ at the layer input grows
$25.9 \to 131.9$ ($5.1\times$) and $\|m_L\|$ grows $8.7 \to 51.1$
($5.9\times$), so the mean's \emph{share} of the stream stays in
$0.33$--$0.44$ throughout; the pre-norm mean grows with the stream. LayerNorm then divides that growth
out, and the object gates read is flat, $\|\bl\| = 6.6$--$10.1$ over
the same span. The resting term (each gate's pre-activation at the mean
input, $w\cdot\bl + b_n$; \S\ref{sec:bias}) inherits the stability
directly (median $-1.57$ at L6 to $-1.19$ at L28, within a band of about
$0.7$ while the stream quintuples). The twin
separates this as a trained property: there the share is not
stabilised at all but climbs monotonically ($0.54 \to 0.78$ over the
same layers) and $\|\bl\|$ climbs with it ($27.2 \to 39.6$). What the network maintains is therefore a constant \emph{fraction}
of the stream: the pre-norm magnitude grows, and the post-norm operating
point is the invariant.

\paragraph{What holds the fraction.} The measurements above show this
stability; an exact decomposition explains it. Writing $x_{\text{ln}} = \bl + \delta$, the
reference component is
$\langle x_{\text{ln}}, \hat{\bl}\rangle = \|\bl\| +
\langle \delta, \hat{\bl}\rangle$, and because $\bl$ is the
corpus mean the second term has mean zero by construction: only its
variance is at issue. Two things suppress it. First, LayerNorm's
per-position denominator: recomputing every contribution with $\sigma$
frozen at its corpus mean raises the variance of the reference
component from $0.62 / 0.51 / 0.96$ to $4.25 / 2.28 / 2.83$ at
L8/L16/L24, so the per-position gain control removes $85\% / 77\% /
66\%$ of it. Second, aggregation: the remainder is a sum over the
$17$--$49$ writes upstream of the layer, each contributing a standard
deviation of at most $0.33$ against a total mean near $6.7$, and the
sum is far steadier than its parts (coefficient of variation $0.118$
against $0.25$--$6.65$ for individual sources). The sources do
not cancel each other: their contributions are weakly positively
correlated (mean pairwise correlation $+0.05$), and covariance adds
$34$--$72\%$ of the total variance. The stability therefore comes from
normalisation and averaging. The same decomposition has two by-products.
The largest and steadiest single contributor is the layer-0 block: its
attention and MLP writes supply $28\%$ of the reference component at L8,
with mean-to-spread ratios of $3.1$ and $4.0$ against $1.3$--$1.5$ for
late writes. And aggregation de-correlates as well as stabilises: a
held-out ridge regression on twelve surface text features predicts the
largest individual contribution at $R^2 = 0.34$ but the total at only
$0.11$ (permutation null $-0.001$), so the total tracks surface text far
less than its parts do.

\paragraph{A slowly rotating band of directions.}
$\cos(\mhat_{10}, \mhat_{18}) = 0.94$, and the reference gates
actually read is aligned over the same span
($\cos(\hat{b}_{10}, \hat{b}_{18}) = 0.92$). Extending the
comparison to every layer shows a band:
layers four apart sit at $\approx 0.97$ (adjacent ones at $0.99$), but
$\cos(\hat{b}_{6}, \hat{b}_{22}) = 0.81$ and
$\cos(\hat{b}_{6}, \hat{b}_{30}) = 0.24$. The coupling survives
the drift: substituting another layer's reference into layer $L$'s resting term, at layer
$L$'s own magnitude, still orders layer $L$'s duty at Spearman
$\rho = 0.86$--$0.96$ anywhere mid-stack and at $0.69$--$0.77$ even
from L30, against $0.06$--$0.09$ for a random direction at matched norm
(\S\ref{sec:bias}; this contrast carries the result, since $\rho$ alone
is near-tautological); in the twin the same transfer instead tracks the
cosine almost exactly ($0.44 \to 0.41$, $0.71 \to 0.70$), so the
tolerance is trained. Causally the same point is made in
\S\ref{sec:identity}: layer 10's gates accept layer 18's mean
direction as their reference almost without loss. This cross-layer
geometry is specific to Phi-2 and varies across the replication
families (Limitations); every
other claim in the paper is per-layer and stands without it.

\paragraph{Content.} Decoded through the \emph{final} unembedding,
mid-stack $\bl$ (L2--L18) aligns only weakly with any single token
direction (max row cosine $0.14$--$0.15$, about $1.5$--$1.7\times$ the isotropic floor), rising toward the readout. This paragraph is
descriptive; no claim in \S\ref{sec:function}--\S\ref{sec:formation}
depends on it.

\paragraph{Sources.} The embedding mean is negligible (norm 0.42,
cosine 0.02 to the deep mean, $m_L$ at L30). Instead, every layer's attention
\emph{and} MLP mean writes are mutually aligned with the deep mean
(cosines $+0.10$ to $+0.77$ through L27, never negative), each small
(norms $\sim$1--4 through L27, against a mean norm of $9$--$51$; the last
four layers write larger and less consistently aligned), summing
coherently: the reference is a distributed construction to which every
component contributes a small aligned constant term. Part of that
constant is input-independent: the module biases and the LayerNorm bias
supply a share that depends on the architecture ($41$--$47\%$ in Phi-2;
\S\ref{sec:bias}).

\section{Function: the gate population thresholds against it}
\label{sec:function}

\paragraph{Manipulation.} At layer $L \in \{6, 10, 14\}$, the residual
entering the layer is modified at all positions $\geq 16$:
$x' = x + (\alpha - 1)(x \cdot \mhat_L)\,\mhat_L$, for
$\alpha \in \{0, 0.5, 1, 1.5, 2\}$. $\alpha{=}1$ is the identity
(verified as an exact no-op: the output distribution is unchanged); $\alpha{=}0$ removes the
stream's projection on the reference; $\alpha{=}2$ doubles it. The
control subtracts the same per-position magnitude along a fixed random
unit direction (norm-matched). Firing is measured as frac(act$>$0.5),
the fraction of post-GELU gate activations above $0.5$, and output damage
as the KL divergence of the final-position next-token distribution from
the clean run's. 39 documents.

\begin{table}[t]
\centering
\small
\begin{tabular}{lrrrrrr}
\toprule
$\alpha$ & \multicolumn{3}{c}{$\Delta$frac(act$>$0.5)} &
\multicolumn{3}{c}{KL vs.\ clean} \\
 & L6 & L10 & L14 & L6 & L10 & L14 \\
\midrule
0.0 (removed) & $+0.103$ & $+0.086$ & $+0.075$ & 1.427 & 0.519 & 0.274 \\
0.5 & $+0.023$ & $+0.023$ & $+0.021$ & 0.047 & 0.040 & 0.035 \\
1.0 (identity) & 0.000 & 0.000 & 0.000 & 0.000 & 0.000 & 0.000 \\
1.5 & $-0.006$ & $-0.007$ & $-0.007$ & 0.014 & 0.018 & 0.020 \\
2.0 (doubled) & $-0.008$ & $-0.009$ & $-0.009$ & 0.055 & 0.060 & 0.110 \\
\midrule
norm-matched control & $+0.0028$ & $+0.0015$ & $+0.0033$ & 0.020 & 0.022 &
0.025 \\
\bottomrule
\end{tabular}
\caption{Dose-response of gate firing and output damage to scaling the
stream's projection on $\mhat_L$ at the manipulated layer (Phi-2,
trained). The twin is flat at every $\alpha$
($|\Delta\text{frac}| \leq 0.003$, no ordering). Point estimates are
document medians over 39 documents; 95\% document-bootstrap CIs (2000
resamples) on the removal row: $\Delta$frac
$[+0.096,+0.112]/[+0.079,+0.098]/[+0.068,+0.088]$ and KL
$[1.14,2.03]/[0.36,0.75]/[0.15,0.40]$ (L6/L10/L14), each disjoint from
the norm-matched control's KL CI ($\leq [0.010,0.032]$). The
document-sampling spread on KL dwarfs the run-to-run fp16 jitter
($\sim$1\%), so only the leading two digits are informative.}
\label{tab:dose}
\end{table}

\paragraph{Result.} Table~\ref{tab:dose} shows the dose-response.
Removal releases firing;
amplification suppresses it; the response is monotone through the
identity point at every layer and $23$--$59\times$ the norm-matched
control, consistent with near-binary gates driven across threshold
together, as reported for GPT-2's MLP units \citep{balogh2026}. Against the clean
baseline of frac(act$>$0.5) $= 0.0105$ at L10, the release is a
$\sim$9$\times$ multiplication of above-threshold firing. The
suppression side saturates small, which is the expected asymmetry: at
duty 0.076 the population is already near its inhibited floor, so
added inhibition has little left to turn off while release frees every
marginal gate. The twin shows no response, having no resting coupling
to lose (\S\ref{sec:bias}).

\paragraph{Output damage.} Removal at a single site costs
KL $= 1.43 / 0.52 / 0.27$ (L6/L10/L14) against $0.020$--$0.025$ for
the norm-matched control. The twin loses only $0.32$--$0.38$ KL under
the same removal despite its raw mean being $8$--$10\times$ larger in norm (121--208 vs.\ 15--21): per unit of removed norm the trained model is
roughly two orders of magnitude more sensitive.

\paragraph{A matched-energy scale.} Two matched-energy controls place
the removal cost. Isotropic noise injected at one layer carrying
the reference's own share of the stream (norm share $0.33$,
\S\ref{sec:object}; about $10\%$ of the energy) costs
KL $0.013$ at both L6 and L10, against $1.68$ (L6) and $0.69$ (L10) for
removing $\bl$ on the same per-position measure (a mean over positions,
distinct from Table~\ref{tab:dose}'s final-position KL): the reference
removal is $127\times$ (L6) to $52\times$ (L10) more damaging than
matched-energy noise, and $670$--$4200\times$ more than removing a random
direction. Turned around, injecting per-position noise \emph{along}
$\hat{\bl}$ costs $8$--$42\times$ the same energy injected along a
matched random direction, across layers and amplitudes; enforcing a
near-zero empirical mean on the injection (residual
$\|\langle n\rangle\|/\|\bl\| \leq 0.01$ in three of four settings,
$0.06$ in the fourth) leaves the ratio unchanged, so the cost comes from
per-position jitter on the reference axis and not from a shift of the
operating point. Both corroborate the direction specificity of
Table~\ref{tab:dose} by an independent route, measured on KL.

\paragraph{Where the damage lands.} Resolving the same removal by
context locates the cost. Applying it at L16 on a 16-sentence corpus
costs $\Delta$NLL $= +0.493$ (change in next-token negative
log-likelihood) and $\Delta H = +0.379$ (change in output entropy) while keeping
top-1 at $0.70$, and the loss is concentrated: $+0.94$ at
high-entropy positions against $+0.04$ at low-entropy ones, and
$+0.88$ at function-word positions against $+0.36$ at content words.
Replacing the massive-activation coordinates with their corpus
constants at the same site is a much milder intervention on the same measure ($\Delta$NLL $= +0.086$, top-1 $0.87$), and its damage sits in
the same places. Removing the reference therefore costs most at the high-entropy, least-certain
predictions and little where the next token is already near-settled.

\paragraph{An input-side demonstration, and what supplies the
reference.} Every manipulation above is an activation edit, which leaves
open the objection that edited states are off the distribution the model
ever visits \citep{mishra2026}. The same conclusion follows from a real input
with no edit: feeding a degenerate input (one token repeated) inflates LayerNorm's per-position denominator by $8$--$15\times$, which
divides the reference out: the reference
component falls from $6.8$ to $0.9$ at L8 and from $7.1$ to $1.0$ at
L24, going slightly negative in the most degenerate condition. Because
the coupling is inhibitory, removing it releases firing, and duty rises
from $0.087$ to $0.791$ at L24 (a $9\times$ increase) with the
weight-space resting term's correlation with measured mean
pre-activation falling from $0.95$ to $0.16$. This is the same
release the $\alpha$ scaling produces, obtained from a real prompt.
Repeated-token input is also known to disrupt the attention-sink
circuit, with a small set of early-layer MLP neurons raising the
hidden-state norm of the repeated tokens \citep{yona2025}; that is
consistent with the inflated LayerNorm denominator measured here, and
the measurement here concerns its effect on the MLP gates.

Two scope statements belong with it, because they bound how much the
model depends on its input distribution. The failure needs extreme
degeneracy: tiling the first $k$ tokens of a document, the denominator
is inflated at $k=1$ and $k=2$ but is already back to its prose value at
$k=4$ ($\sigma = 15.2 / 2.7 / 0.92$ against a prose $0.99$), and an
i.i.d.\ draw from four distinct token types is enough. And recovery is
local: appending a clean prompt after a degenerate context brings duty
most of the way back within two or three tokens ($0.739 \to 0.542 \to 0.207 \to 0.186$ by
offset at L24, against a prose $0.08$). Natural text is therefore
nowhere near the failure regime, and a short window of ordinary recent
context is enough to restore the reference. 

\paragraph{Reach in sequence position.} A single-site removal is local
in depth (\S\ref{sec:maintenance}) but reaches far along the sequence:
removing the reference in an eight-position
window at L6 and then reading only \emph{untouched} later positions, the
next-token distribution is still altered $32$ tokens past the window at
KL $= 0.098$, $16\%$ of the in-window magnitude and $54\times$ the
norm-matched control, and the ratio to control \emph{grows} with
distance, since the control decays faster than the effect. Greedy
continuations diverge from the unperturbed ones within a median of two
steps. The effect alters later predictions without making them worse on
average: KL is large, but the change in the true token's log-probability
is inconsistent in sign.

\paragraph{Cross-model.} The same manipulation at one mid-stack layer (depth
$\approx 0.31$: Pythia-1.4B $L8$, Qwen2.5-1.5B $L9$; 29 documents,
each against its own twin) reproduces the sign, the monotonicity, and
the control separation. On the duty-cycle measure (activation $>0$),
removal shifts duty by $+0.318\,[0.303,0.329]$ (Pythia) and
$+0.302\,[0.286,0.316]$ (Qwen), doubling by $-0.124$ and $-0.081$,
monotone across a 15-point $\alpha$ grid, against $+0.0001$ for the
norm-matched control and flat twins ($|\Delta\text{duty}| \leq
0.002$); Phi-2 on the same measure is $+0.238/-0.070$. Qwen is the
sharpest case: its SwiGLU gate has no bias parameter, so the carried
reference is the entire resting inhibition, and scaling it still moves
firing in dose-response. Removal costs final-position KL $3.60$
(Pythia) and $0.59$ (Qwen). Weakening the reference also raises output
entropy in all three models tested ($\Delta H$ at
$\alpha{=}0.8$ is $+0.057$ Pythia, $+0.047$ Qwen, $+0.013$ Phi-2,
each above a norm-matched control with $|\Delta H| \leq 0.0005$), with the minimum
at the natural magnitude in Phi-2 and Pythia (in Qwen the amplification
side is noisy and dips below it).

\section{Identity: the direction, not the magnitude}
\label{sec:identity}

Removal tests necessity of the projection, but it leaves two
alternative readings open. Gates might not read this direction at all
and simply need the stream to be full; and since the manipulation acts
on the pre-norm residual, removing the projection shrinks the norm
entering LayerNorm, and the renormalisation then boosts the gain on every
orthogonal component; some of the release could come from that gain
boost. The replacement panel
addresses both: at L10,
$x' = x - (x\cdot\mhat_{10})\mhat_{10} + (x\cdot\mhat_{10})\,\hat{c}$ transfers the
signed per-position projection onto a candidate direction $\hat{c}$;
magnitude, sign pattern, and timing are preserved, the total norm is
approximately unchanged (the cross-terms with $\hat{c}$ are small),
and only the direction changes.

\begin{table}[t]
\centering
\small
\begin{tabular}{lrrr}
\toprule
replacement $\hat{c}$ & $\cos(\hat{c},\mhat_{10})$ &
$\Delta$frac(act$>$0.5) & KL \\
\midrule
none (pure removal) & --- & $+0.0845$ & 0.516 \\
coordinate-shuffled $\mhat_{10}$ & $-0.003$ & $+0.0840$ & 0.950 \\
random direction & $+0.011$ & $+0.0778$ & 0.484 \\
function-word $W_U$ sum & $+0.096$ & $+0.0740$ & 0.395 \\
layer 18's $\mhat_{18}$ & $+0.940$ & $+0.0027$ & 0.004 \\
identity & 1.000 & 0.000 & 0.000 \\
\bottomrule
\end{tabular}
\caption{Replacement panel at L10 (Phi-2, 23 documents). Preservation
of the baseline firing level tracks $\cos(\hat{c},\mhat_{10})$ monotonically.}
\label{tab:identity}
\end{table}

In Table~\ref{tab:identity}, shuffled and random directions release
firing almost exactly like removal; a direction at cosine 0.096 to $\mhat_{10}$
buys a sliver proportionate to its cosine and nothing more; and
another layer's mean direction (cosine 0.94) preserves baseline firing almost perfectly, $31\times$ less release and $130\times$ less KL than
removal. The magnitude-reader alternative is excluded: no
low-cosine candidate preserves anything. The gain-boost alternative is
excluded the same way: a random-direction replacement leaves the
pre-norm magnitude in place, so the LayerNorm gain is the same as
clean, yet firing releases as if the projection had been removed,
while layer 18's direction, under the same norm, preserves baseline firing. One further observation: the
shuffled replacement's KL (0.95) \emph{exceeds} removal's (0.52);
transferring the projection onto a wrong direction injects a coherent
error signal on top of losing the reference.

\section{Implementation: the threshold is the coupling}
\label{sec:bias}

For a gate $n$ with read row $w$ (a row of Phi-2's \texttt{fc1}) and
bias parameter $b_n$, the \emph{resting} pre-activation decomposes
exactly:
$\langle w\cdot x_{\text{ln}} + b_n \rangle = w\cdot\bl + b_n$, since
$\langle x_{\text{ln}} \rangle = \bl$ by definition. If gates
threshold against the carried reference, training should have moved
the operating-point inhibition into the coupling $w\cdot\bl$ rather
than the explicit bias parameter $b_n$.

\begin{table}[t]
\centering
\small
\begin{tabular}{lrrrrrr}
\toprule
layer & $w\cdot\bl$ (median) & $b_n$ (median) & resting &
ref-dominant & $\rho$(resting, duty) & $\rho$(random dir) \\
\midrule
6 & $-1.535$ & $-0.027$ & $-1.568$ & 99.98\% & 0.943 & 0.058 \\
10 & $-1.226$ & $-0.022$ & $-1.250$ & 99.99\% & 0.959 & 0.072 \\
14 & $-1.091$ & $-0.021$ & $-1.115$ & 99.96\% & 0.960 & 0.085 \\
22 & $-1.130$ & $-0.018$ & $-1.149$ & 99.94\% & 0.954 & 0.073 \\
\bottomrule
\end{tabular}
\caption{Exact resting decomposition of gate pre-activations (Phi-2,
19 documents). ``Ref-dominant'' is the fraction of gates whose resting
inhibition is carried mostly by $w\cdot\bl$ ($|w\cdot\bl| > |b_n|$). The last column repeats
the rank correlation with a random direction substituted for $\bl$ at
matched norm (median of 50 draws, $p_{95} \leq 0.192$, maximum
$0.223$; from a 39-document run, which reproduces the
$\rho$(resting, duty) column at $0.946$--$0.964$). The
$\rho$(resting, duty) column is near-tautological read alone; it is
the \emph{contrast} between the last two columns that carries
direction specificity. 95\% document-bootstrap
CIs (2000 resamples): ref-dominant stays $\geq 99.89\%$ at every layer
(widest $[99.89,99.98]$ at L22), and $\rho$ is
$[0.936,0.945]/[0.951,0.959]/[0.953,0.959]/[0.945,0.952]$
(L6/L10/L14/L22; percentile intervals, and resampling biases $\rho$
slightly low, so a point estimate can sit just above its interval). The
$48$--$56\times$ quoted in the text is the ratio of median magnitudes,
$|w\cdot\bl|$ against $|b_n|$ (the latter $0.027 / 0.023 / 0.023 /
0.021$), which differ slightly from the signed medians shown. The
\emph{resting} column is a per-gate median and
does not equal the sum of the $w\cdot\bl$ and $b_n$ medians, since a
median does not distribute over addition across a heterogeneous gate
population.}
\label{tab:bias}
\end{table}

Table~\ref{tab:bias}: the reference coupling carries the resting
inhibition at $48$--$56\times$ the explicit bias parameter (ratio of
median magnitudes) for more than 99.9\% of gates at every tabulated layer
(the shallowest layer measured, L2, is at 98.5\%), and the resting
pre-activation rank-predicts each gate's duty cycle at
$\rho \approx 0.95$. The $\rho \approx 0.95$ is near-tautological
on its own, and the next paragraph derives its null; what survives is
the narrower claim that the population's operating point is set along
this direction and not by the bias parameter. In the twin,
$w\cdot\bl$ has no shared sign (median $\leq 0.01$, half the gates on
each side, typical magnitude $0.5$) and duty is $\approx 0.5$; its bias
parameters are exactly zero at init, so the informative twin statistic is
that the population has no shared resting inhibition. This is the
classical bias-as-weight-to-a-constant-input construction, discovered
in the weights: the model implements thresholds as couplings to a
constant it itself carries. It also explains the
dose-response of \S\ref{sec:function} without further assumptions:
scaling the stream's projection on $\mhat_L$ scales the operating-point
term of every gate simultaneously.

\paragraph{What the duty correlation does and does not show.} Since
$\bl := \langle x_{\text{ln}} \rangle$, the resting term is the
mean pre-activation, and duty is the fraction of that same
distribution above zero. A rank correlation between the mean of a
distribution and $P(>0)$ is high for any family with comparable
spreads, so $\rho \approx 0.95$ may be a property of the construction
rather than of $\bl$, and the twin cannot adjudicate: with resting
$\approx 0$ for every gate it has no spread to correlate. We therefore
built the null the twin does not provide. Permuting the resting values across gates (each gate keeps its own centred pre-activation
distribution but receives an offset carrying no information about
$\bl$) still yields $\rho = 0.87$--$0.93$ (L6/L10/L14/L22), which is
$92$--$97\%$ of the measured value. Injecting offsets drawn from the
trained resting distribution into the \emph{twin}, a model with no
coupling at all, yields $\rho = 0.984$, higher than the trained model,
because the statistic is governed by how homogeneous the per-gate
spreads are (twin coefficient of variation $0.056$ against the trained
model's $0.15$--$0.20$) rather than by the reference. Consistently, with each gate's resting term divided by its own
pre-activation standard deviation $\sigma_n$,
$\rho(\text{resting}/\sigma_n, \text{duty}) = 0.987$--$0.994$ exceeds
the raw figure. The duty correlation is a location fact and we do not
rest anything on it.

Two statistics do survive that test. First, direction specificity: a
random direction at the same norm as $\bl$, substituted into the same
decomposition, orders duty at $\rho = 0.058$--$0.085$ (median over 50
draws; $p_{95} \leq 0.192$, maximum $0.223$) against the reference's
$0.946$--$0.964$. The operating point is set along this specific
direction, the static counterpart of the replacement panel of \S\ref{sec:identity}. Second, the coupling is
largely norm-scaled: the read-weight norm alone orders duty at
$\rho = -0.878$ to $-0.896$, as the per-gate anatomy below implies (the coupling is a
narrow, nearly fixed fraction of each row's norm), and in the twin the same statistic is null ($-0.010$ to
$+0.007$), so the norm-to-duty relation is itself trained. The
read-weight norm accounts for most of the duty ordering; what $\bl$
contributes is the part left after that norm-only ordering.

\paragraph{Per-gate anatomy.} Factoring each fc1 row against the unit
reference, $w = c_n\,\hat{\bl} + w_\perp$, the coupling is a small
slice of read norm: $|c_n|/\|w\|$ has median $0.07$--$0.12$ across
layers against a random-twin floor of $0.013$, thinning toward the
readout, so $\approx 0.99$ of the row remains free for content. The
coupling alone rank-predicts duty as well as the full resting
inhibition ($\rho(c_n,\text{duty}) = 0.943$--$0.959$, matching
Table~\ref{tab:bias}), so $b_n$ adds nothing to the ordering; both
figures inherit the caveat above. On its own $b_n$ also orders
duty ($\rho = 0.81$--$0.83$), but it adds nothing once the coupling is
known. Per gate, the coupling takes $0.5$--$1.4\%$ of the read weight's
energy and fixes where the gate sits while leaving its content direction
almost untouched; $c_n = w\cdot\hat{\bl}$ is both the gate's (negative)
read of the reference and the term that sets its resting operating point.

\paragraph{The coupling over training.} The decomposition above
compares two endpoints. Pythia-410M-deduped, whose GELU gates carry an
explicit bias and whose checkpoints are public, lets the two static
terms be tracked step by step (four mid-stack layers). $b_n$ is
exactly zero at initialisation (Pythia zero-inits biases) and stays
$\leq 0.04$ in magnitude the entire run, while $|w\cdot\bl|$, after
an initial drop ($\sim0.25 \to 0.12$ by step 128), builds through the step $512\to1000$ window ($0.16 \to 0.38$),
reaches $0.84$--$0.95$ by step 4000, peaks at step 8000 (16000 at
the shallowest layer), and relaxes partially thereafter. The reference coupling carries the resting inhibition of
$\geq 99.6\%$ of gates at every checkpoint after initialisation, where $b_n$ is exactly zero: there is no
phase in which the explicit bias leads and is overtaken. The network
does not route the operating point through the bias parameter at any
stage; it builds the coupling directly.
Duty falls
$0.51 \to 0.16$ over the same build and $\rho(\text{resting},
\text{duty})$ stays $0.91$--$0.998$ throughout. The time-course confirms
the endpoint comparison, and it places the onset in the same window as
\S\ref{sec:formation} through a different statistic.

\paragraph{Cross-model replication.} The resting decomposition is
static (weights and mean activations only), so it can be re-run
cheaply on other families against their twins.
Table~\ref{tab:bias_xmodel} does so on four more architectures,
spanning both gate types: GELU MLPs with an explicit gate bias
(GPT-2-medium, Pythia-1.4B) and bias-free SwiGLU gates (Qwen2.5-1.5B,
TinyLlama-1.1B). In every family the resting pre-activation is
negative (gates held off by the carried reference) and rank-predicts
duty at $\rho \geq 0.91$ at every sampled layer but one (below), with the twin
floored (resting $\approx 0$, duty $\approx 0.5$) throughout. Every
$\rho$ in this paragraph and in Table~\ref{tab:bias_xmodel} inherits
the caveat above and is reported as description; what carries the
cross-family claim is the sign and magnitude of the resting term
relative to $b_n$, and the random-direction control was run on Phi-2
only. Where a
gate bias exists, the reference coupling dominates it for $>99\%$ of
gates at $15$--$43\times$ the bias parameter; where the gate is
bias-free, the coupling alone carries the entire negative resting
term. The one sub-threshold $\rho$ is Qwen's shallowest sampled layer
($L5$, $\rho = 0.62$, a strongly suppressed near-off regime, duty
$0.004$); every deeper layer is $\rho \geq 0.96$. Magnitude varies
(TinyLlama's resting inhibition is an order of magnitude weaker than
GPT-2's).

\begin{table}[t]
\centering
\small
\begin{tabular}{llrrr}
\toprule
model & gate (bias?) & resting med.\ & ref-dom.\ & $\rho$(rest, duty) \\
\midrule
GPT-2-medium   & \texttt{c\_fc} (yes)         & $-1.52$ to $-0.99$ & $99.4$--$99.7\%$ & $0.91$--$0.97$ \\
Pythia-1.4B    & \texttt{h\_to\_4h} (yes)     & $-1.08$ to $-0.74$ & $>99.9\%$        & $0.96$--$0.98$ \\
Qwen2.5-1.5B   & \texttt{gate\_proj} (no)     & $-4.14$ to $-0.37$ & ---              & $0.62$--$0.99$ \\
TinyLlama-1.1B & \texttt{gate\_proj} (no)     & $-0.21$ to $-0.07$ & ---              & $0.97$--$0.99$ \\
\bottomrule
\end{tabular}
\caption{Cross-model resting decomposition (30 documents, four
mid-stack layers per model at depth fractions $0.2$--$0.65$; ranges
span those layers). ``Ref-dom.'' is the fraction of gates whose
resting inhibition is carried mostly by $w\cdot\bl$ rather than the
bias parameter (vacuous for the bias-free SwiGLU gates; for these the
informative statistic is the negative resting coupling and its
duty-prediction). Where a bias exists, $|w\cdot\bl|/|b_n|$ is
$15$--$18\times$ (GPT-2) and $38$--$43\times$ (Pythia). Every model's
random-init twin is floored: $|\text{resting median}| \leq 0.01$, duty
$\approx 0.5$.}
\label{tab:bias_xmodel}
\end{table}

\paragraph{A counteracting boundary case.} Extended across the
eight-model GELU/ReLU/SiLU scan, the static decomposition finds the
mechanism in all six non-OPT models (five families; ref-dominant $99.4$--$100\%$ where a bias exists,
$\rho \geq 0.91$ except at Qwen's shallowest layer) and not in the two OPT models (opt-350m,
opt-1.3b): their resting inhibition is $40$--$130\times$ weaker,
bias-dominated, and rank-predicts nothing ($\rho = 0.000$, as follows
from a resting term near zero). A norm
decomposition explains why. Writing $w\cdot\bl = w\cdot(\bl - \beta) +
w\cdot\beta$ (residual-carried mean plus LayerNorm bias), the
residual-carried part is primary wherever the mechanism is present:
it dominates $91$--$99\%$ of gates in the GELU LayerNorm models and
equals the resting term by construction in the bias-free RMSNorm
models, with $\beta$ a $7$--$26\%$ augmentation. So against the LayerNorm
bias, the \S\ref{sec:object} attribution of the reference to a
residual-carried mean holds; the module biases are separated next. OPT shows what
the mechanism requires: it builds a large
residual-carried inhibition comparable to GPT-2's but trains an almost
exactly opposing LayerNorm-bias excitation ($|w\cdot(\bl-\beta)| \approx
|w\cdot\beta| \approx 0.5$--$0.6$, opposite sign), netting
$w\cdot\bl \approx 0.006$: the reference is present but cancelled at
the operating point. ReLU-versus-family stays formally confounded (every
ReLU model sampled is an OPT).

\paragraph{How much of the reference is a parameter.} The decomposition
above separates the residual-carried mean from the LayerNorm bias,
but it counts the accumulated \emph{module} biases (each block's
attention-output and MLP-output bias) inside the residual-carried
part, though they are input-independent, added unconditionally whatever
the input. Separating them changes the accounting materially. Projecting
$\beta$ and every upstream module bias onto the reference (the module
biases enter as a constant vector divided by the per-position $\sigma$,
so the constant is computable from weights alone), the input-independent
share of the reference component tracks architecture exactly across the
replication set: $41$--$47\%$ (Phi-2), $10$--$32\%$ (GPT-2-medium), $52$--$84\%$ (Pythia-410M),
$26$--$40\%$ (Pythia-1.4B), and \emph{exactly} $0\%$ in all three
bias-free RMSNorm models (Qwen2.5-1.5B, TinyLlama, SmolLM2), which have
neither $\beta$ nor output biases and must therefore assemble the same
reference entirely from computed writes. OPT is again the informative
case: its share is \emph{negative} ($-16$ to $-8\%$ in opt-350m), the same
cancellation seen in the paragraph above from the parameter side. So the
mechanism holds across the non-OPT architectures while its
\emph{ingredients} vary: how much of the reference is carried in the stream versus in the
parameters is a proportion the architecture sets, and the
span is wide, so no single figure should be quoted for it.

\section{How the reference persists, and the cost of losing it}
\label{sec:maintenance}

\paragraph{Attention carries the steady feed.} Decomposing each
block's output into its component along $\mhat_L$ and the orthogonal
remainder, and measuring the constant share of each (the product of its means over two
halves of the corpus, as a fraction of its mean square): at
L10, the attention block's $\mhat_L$ channel has constant share 0.452
against 0.014 for its orthogonal channel, a $33\times$ dissociation;
the block outputs a steady reference feed while its content channel
fluctuates. The MLP shows no such dissociation (0.003 vs.\ 0.003): unlike
attention it carries no steady reference feed, though its time-averaged
write is reference-aligned (\S\ref{sec:object}). In the twin both channels are largely constant (0.96
vs.\ 0.86; an untrained block is a mostly-constant filter), so the
dissociation is a trained property. Together with \S\ref{sec:object}:
per-layer mean writes build the reference; attention sustains it in
operation.

\paragraph{Restoration.} A single-site removal is local in depth:
downstream firing is undisturbed. After removal at L6
(\S\ref{sec:function}), $\cos(\text{resid}, \mhat_L)$ falls by $0.16$ at
L10 and L14 and by $0.10$ at L18, and recovers to baseline by L26--L30: the
stream re-accumulates its mean over $\sim$12--16 layers, consistent with
the sources of \S\ref{sec:object} (every layer's mean write is
reference-aligned, so the reference restores itself by accumulation).
Deeper removal sites restore more slowly.

\paragraph{Co-removal.} Single-site removal understates the
dependence because the stream heals. Removing the projection at
\emph{all} measured layers $\geq 6$ simultaneously releases firing at
every depth (most at the shallowest site, still $+0.075$ at L18) and costs
KL $= 4.38$, $\Delta$NLL $= +4.75$ nats, about $8.4\times$ the
L10 single-site removal (Table~\ref{tab:dose}, KL $0.52$), while the
twin under the identical intervention
loses 0.65 KL with flat firing (document-bootstrap 95\% CIs over 23
documents: KL $[3.74,6.05]$, $\Delta$NLL $[4.62,5.03]$, disjoint from
the twin). Late-only co-removal ($\geq 14$) is far milder (KL 0.55,
CI $[0.24,0.90]$): the dependence is concentrated mid-stack. The
reference is a standing condition across the middle of the network, and
no single layer carries the dependence alone.

\section{Development}
\label{sec:formation}

The operating point of \S\ref{sec:bias} is built early in training, and
by a token-local pressure. Three developmental settings, at increasing
resolution, show this.

\paragraph{Pythia-410M: the reference forms in the induction window.}
On Pythia-410M-deduped public checkpoints (a different family and scale
from the causal experiments; one seed), the reference coupling forms in
the same early window as gate sparsification and the induction
transition. In the step 512$\to$1000 interval the induction-head score
rises $0.03 \to 0.88$ \citep{olsson2022}, the median duty cycle falls
$0.42 \to 0.28$ (continuing to $0.16$ by step 4000), and the reference
share of drive deepens $-0.06 \to -0.34$ (to $\approx -0.75$ by
step 3000--4000); the static coupling $|w\cdot\bl|$ builds in the same
window (\S\ref{sec:bias}). Pythia has no checkpoint between 512 and 1000, so this
grid places the onset in that one interval but cannot order the three
within it. Below it nothing is yet moving: across log-spaced steps
1--256 the induction score is at its floor, duty $\approx 0.50$, and the
reference share $\approx 0$ at mid-stack. The only earlier event is
training's first act on the MLP write, which destroys the constant the MLP writes at
initialisation (the write's constant share, as in \S\ref{sec:maintenance},
falls $0.40 \to 0.15$ by step 128). Massive activations follow a
different early course in the same model family: absent at
initialisation, they rise to an early peak in shallow and deep layers
\citep{gallego2025}. The trained reference is therefore built during
training, and it is in place by the time the induction transition
completes.

\paragraph{A 160M model: the operating point forms first.} A finer
developmental grid resolves the order the Pythia checkpoints cannot. In
a 160M model (Pythia-160M architecture, a standard GELU MLP, trained
here on the Pile with dense early checkpoints), the gate operating point
forms well before the induction transition: median mid-stack duty falls
from $0.50$ to $0.075$ by step $\approx 800$ (and lower thereafter), and
the resting depth $z$ (the resting pre-activation divided by its
per-unit standard deviation $\sigma_n$, which puts the operating point in
units of the gate's own spread) reaches $\approx 1.5$ over the same window, while
the induction-head score is still near its floor and completes its rise
only over steps $\approx 3000$--$12000$. Throughout, the operating point is the reference
coupling (the mean-input term carries $\approx 0.99$ of the resting
pre-activation, as in \S\ref{sec:bias}), so it is the mechanism of
\S\ref{sec:bias} that is in place first. (Absolute step counts are not
comparable to Pythia-410M's, which uses a different schedule and
$\sim$100$\times$ more tokens per step; the developmental \emph{order}
is the point.)

\paragraph{An attention-free model: the operating point is token-local.}
The operating point forms without any cross-token computation. Training
the same 160M architecture with attention disabled (every layer's
attention output projection zeroed and frozen, leaving a purely
position-wise model that predicts each token from the current token
alone, a bigram map) reproduces the same development. In a matched pair of runs differing
only in whether attention is enabled (identical script and schedule, to
step 3000), the attention-free model's mid-stack duty settles at $0.077$
against $0.067$ for the attention-carrying one (a $\approx 15\%$ higher
floor) and its operating point is slightly shallower ($z \approx 1.3$
against $\approx 1.5$), while the reference coupling is unchanged
(the mean-input term carries $\approx 0.99$ of the resting pre-activation in both). Attention makes
sparsification marginally faster and deeper, but the operating point forms under the token-local next-token
prediction task alone, whether or not the model can read other positions
(one seed per arm).

\paragraph{Reading.} Across the three settings (each one seed) the
operating point is an
early, token-local component of a trained MLP: the network sets its
gates against the carried reference as it learns the base next-token
map, in the same early window in which cross-token circuits mature at
scale. Activation sparsity is a known endpoint property
\citep{li2023lazy,szatkowski2025}, and \citet{shvetsov2026} give an
optimisation account of early training that drives weights negative
(shown in GPT-nano, for weights that read positive activations); the
reference coupling is how that endpoint is reached here, and the pressure that reaches it is present even in a model
with no context to read.

\section{Discussion}

Representational analyses commonly subtract the corpus mean, and for many
questions that is the right step: mean-centering for analysis leaves the
model's computation untouched. The results here say what the subtracted
component is. It is the reference against which the gate population's
default-off state is defined; removing it from the forward pass costs KL
$0.27$--$1.43$ at a single site in Phi-2 and $4.38$ when it is removed at
every measured layer from L6 on, and the damage falls on the least
certain predictions (one site, 16 sentences). An account of gate
behaviour built on mean-centred activations has set aside the term that
decides where the gates rest, and needs it back when the question is why
a unit is off. The same mechanism appears in sparse autoencoders: a
feature's encoder pre-activation is shifted by its alignment with the
activation mean, so features anti-aligned with it start with negative
pre-activations and can stay dead, and centering the autoencoder's input
removes the effect \citep{simon2026}; there the effect is large in
vision and protein models and small in the language models tested. On
the architecture side, PaLM dropped biases from its
dense kernels and layer norms for training stability
\citep{chowdhery2022}; the decomposition suggests why this need not cost
anything on the gate side, since outside OPT the operating point never
lived in $b_n$: the network builds the coupling from early training even
when a bias is available (\S\ref{sec:bias}, the coupling over training), and the bias-free RMSNorm
families assemble the whole reference from computed writes
(\S\ref{sec:bias}).

The coupling can also be followed into MLPs built from gated linear
units (GLU), where the read
stage has a gate branch and a value branch. There it sits in the gate
branch, and a gate unit's coupling to $\bl$ places it on a continuum
from a thresholded switch to a bilinear multiplier of its two reads; that
analysis is reported separately \citep{tuomi2026backgroundglu}.

\section{Related work}

\paragraph{Massive activations, sinks, outlier dimensions.} Attention
sinks are initial tokens that receive strong attention regardless of
their content \citep{xiao2023}, and massive activations are near-constant
residual components that \citet{sun2024} describe as ``indispensable bias
terms'', acting as an implicit bias in self-attention; \citet{owen2025}
find that this does not hold uniformly across architectures, since
suppressing massive activations is not harmful in every model.
\citet{chen2026spikes} show that spike-carrying tokens become constant
vectors after normalisation that act as structural biases inside
attention, and later work ties sinks to compression valleys
\citep{queipo2025}, to outlier-driven rescaling essential for training
\citep{qiu2026}, and to the damping of transient amplification across
depth \citep{li2026}. Most recently, \citet{ssun2026} give a functional
split: massive activations as near-constant hidden components acting as
implicit parameters, and attention sinks as local attention modulators;
they identify the pre-norm configuration as what couples the two,
decoupling them under ablation. Outlier dimensions track token frequency
\citep{puccetti2022}, go with high-magnitude LayerNorm scaling parameters
\citep{kovaleva2021}, and can dominate cosine similarity
\citep{timkey2021},
and \citet{macocco2025} show that last-layer outlier dimensions act as
a constant positive logit contribution toward frequent tokens. The
object studied here is in the same family but the measured function
differs in site and form: a \emph{direction} (not a coordinate set),
read at the MLP gate inputs mid-stack (not at the output logits), that
carries the resting inhibition of the gate population by exact
decomposition, is direction-specific under replacement, and is
assembled by distributed aligned writes. The context-resolved removal
of \S\ref{sec:function} does place the two objects in overlapping
\emph{functional} territory, since removing the reference costs most at
function-word and high-entropy positions; we take the separation to be one of
object and site rather than of function, and the geometric comparisons
below are what carry it.

We ran the geometric comparison against these objects on Phi-2.
Mid-stack, $\cos(\bl, \cdot)$ against the first-token direction at its
emergence layer (located here at layer 21, peak coordinate magnitude
$\sim$1250) is $0.04$--$0.17$, and against the residual-sink and
massive-coordinate directions $\leq 0.27$: the objects are distinct.
\citet{shi2026} note that such a first-token massive activation ``can
serve as a stable and shared global reference vector'' for attention,
but argue that its rigidity limits input-dependent attention.
The separation is trained: in the random-init twin the reference
is the attention-sink direction ($\cos = 0.99$, no massive activation), and
training rotates the reference off the sink. On outlier dimensions:
$\bl$ carries $10$--$23\%$ of its squared norm on the twenty largest
last-layer outlier coordinates against an isotropic share of $0.8\%$,
a real but minority overlap that grows toward the readout; its own
dominant coordinates are largely disjoint from the outlier set (top-20
Jaccard $0.05$--$0.14$). Mid-stack, $\bl$ is a distributed direction
with a minority frequency-correlated component, distinct from the
axis-aligned outlier set, and it should not be conflated with the sink objects of
\citet{qiu2026} or \citet{shi2026}. This distinctness is a property of
Phi-2. In models with strong massive activations the mean concentrates
on the massive coordinates: in Qwen3-architecture models trained from
scratch, \citet{cao2026meanbias} find that the dominant activation
outliers late in training are largely induced by a rank-one mean bias,
and in the SwiGLU models of \citet{tuomi2026backgroundglu} the
reference's largest coordinates coincide with the massive-activation
coordinates.

\paragraph{Distinctness under a variance-aware metric.} The comparisons
above use Euclidean cosines, and \citet{ying2026} note that the Euclidean
cosine treats all dimensions equally, although only the dimensions along
which the data vary matter, and a few such dimensions can dominate it
\citep{timkey2021}. That objection is pointed here, because the objects
we compare against (the residual sink and the massive activations)
are those high-variance coordinates, so an
equal-weight cosine could be understating an alignment with them. We
therefore recomputed the comparisons under variance-aware metrics, on
the same directions and corpus, so that any change is attributable to
the metric. In the space $\bl$
is defined in and gates read (the covariance of the layer-normalised
MLP input), the covariance-weighted cosine of \citet{ying2026} places
$\bl$ against the sink, first-token and massive directions at $-0.23$
to $+0.01$, against an empirical random-pair null of $0.03$ median and
$0.11$ at the 95th percentile; restricting to the top-512 principal
subspace gives $-0.41$ to $-0.14$. The positive control, the covariance-weighted
$\cos(\bl, m_L)$, is $0.91$--$0.95$ throughout. Under the
metric the objection prescribes, the objects show no alignment, and
some are mildly opposed. The instrument is validated on the twin, where
$\bl$ is the attention-sink direction: there every metric returns
$0.83$--$0.99$. Both metrics show one trend: the overlap with the sink grows toward the readout
($0.12 \to 0.27$ Euclidean over L6--L18, $0.08 \to 0.33$ in the
principal subspace of the raw-residual covariance), in step with the outlier-coordinate mass
reported above.

\paragraph{Bias as coupling.} Treating bias as a weight onto a
constant input is the classical homogeneous-coordinates construction.
A weight-based taxonomy of gated neurons \citep{gerstner2025} and an
activation-based reading of MLP units as binary routers
\citep{balogh2026} characterise what individual gates do, and massive
activations have been read as carried constants with a bias-like role
\citep{sun2024,ssun2026}. The observation here joins these: a trained
transformer implements the homogeneous-coordinates construction
spontaneously, with the constant supplied by the network's own
carried mean rather than by a dedicated input.
\citet{sun2024} also report the trained converse on the attention
side: augmenting self-attention with explicit learnable bias
parameters yields GPT-2 models that do not form massive activations
(the same remedy is ineffective in some other architectures;
\citealp{owen2025}): given the parameter, the network uses it instead of building a carried
constant. OPT plays the same role on the gate side here
(\S\ref{sec:bias}): the parameter-based strategy exists, and most
families do not take it.

\section{Limitations}

\begin{itemize}
\item \textbf{Replication scope.} The static decomposition
  (\S\ref{sec:bias}) is shown on the eight-model scan and is absent in
  OPT, where an opposing LayerNorm bias cancels the carried reference;
  the ReLU-versus-OPT-family confound (every sampled ReLU model is an
  OPT) is not separated by matched-nonlinearity twins. The causal
  dose-response and the entropy release side replicate on two further
  families (Pythia-1.4B, Qwen2.5-1.5B); the replacement panel
  (\S\ref{sec:identity}, one site, L10) and co-removal
  (\S\ref{sec:maintenance}) are Phi-2-only. The development results
  (\S\ref{sec:formation}) are trend-level and single-seed: the
  Pythia-410M checkpoint time-course is one seed (15--30 documents), and
  the finer-grid and attention-disabled results are each one 160M model
  trained here, one seed per arm. Two further results are single-model
  and modest-$n$: the depth profile of the mean's share of the
  stream (\S\ref{sec:object}) and the context-resolved removal
  (\S\ref{sec:function}, one site, 16 sentences).
\item \textbf{Cross-layer geometry is family-dependent.} The
  rotating-band description (\S\ref{sec:object}), and the transfer
  measurement that goes with it, are Phi-2 only; the cross-layer
  cosine is heterogeneous across the replication families: $\cos(\mhat_L, \mhat_{L'})$
  between mid-stack layers is high in GPT-2-medium ($0.95$) and
  Qwen2.5-1.5B ($0.96$), modest in TinyLlama-1.1B ($0.74$), and below
  its own twin floor in Pythia-1.4B ($0.41$ vs.\ twin $0.67$). The
  static coupling replicates in every non-OPT family and the causal
  dose-response in all three models tested; the single-carried-direction
  geometry does not replicate, and the paper does not claim it.
\item \textbf{Which statistics carry which claim.} The rank
  correlation between the resting term and duty is near-tautological
  on its own (\S\ref{sec:bias}), because the resting term is the mean
  pre-activation and duty is $P(>0)$ of the same distribution; we
  report its nulls and rest nothing on it. What the static
  decomposition establishes is the sign and magnitude of the resting
  term against $b_n$, and, on Phi-2, direction specificity against a
  matched-norm random control. The functional claims rest on the
  causal tests (\S\ref{sec:function}, \S\ref{sec:identity},
  \S\ref{sec:maintenance}), which do not use the correlation at all.
  The duty ordering is also largely norm-scaled ($\rho = -0.89$ from
  the read norm alone), so the reference's contribution to the ranking
  is the residue above that, which we have not decomposed further.
\item \textbf{Within-layer atomicity.} $\bl$ is a corpus mean, rank-1
  by construction; the token-class structure it hides
  (\S\ref{sec:object}) carries only $2$--$4\%$ of the variance around it,
  so the rank-1 treatment is a good approximation for the operating-point
  claims, and every causal result here concerns only the rank-1 mean.
\item \textbf{Operating-point asymmetry.} The entropy response to
  weakening the reference (\S\ref{sec:function}) rests mainly on the
  release direction, since the amplification side is shallow in Phi-2
  and varies across families.
\item \textbf{Manipulation scope.} The reference is the corpus mean; a
  position-resolved reference is a refinement not tested.
  Manipulations leave positions $<16$ untouched, so attention
  re-import from the sink region is possible. $\alpha$-scaling changes
  the LayerNorm input distribution, so all downstream effects pass
  through LayerNorm renormalisation; the norm-matched control shares this
  property, and the norm-preserving replacement panel
  (\S\ref{sec:identity}) excludes a pure gain-boost account at L10,
  the one site where it was run. The function-word replacement is one
  naive construction of a content-direction control.
\item \textbf{Geometric comparisons are activation-proxy.} The
  comparisons against first-token massive activations, the residual
  sink, and last-layer outlier dimensions (Related work) use
  activation-based proxies of each object on one model; a weights-only
  unification against each paper's exact construction is not
  attempted. Two caveats attach to the variance-aware recomputation
  reported alongside them. Weighting instead by the \emph{raw
  residual} covariance is uninformative here, because the comparison
  objects are themselves the high-variance coordinates: that metric
  returns nearly the same value for all three proxies (within $0.004$), and
  its own random-pair null reaches $0.52$--$0.73$ at the 95th
  percentile, so it cannot discriminate. And an inverse-covariance
  (whitening) variant fails its own positive control at this sample
  size ($8{,}868$ positions for a $2560$-dimensional covariance),
  ranking $\cos(\bl,\text{sink})$ above $\cos(\bl, m_L)$ at the weakest
  regularisation (at L6 and L18) and holding the positive control to only
  $0.70$--$0.80$ at the others; we report it
  as excluded. The recomputation is Phi-2-only,
  like the comparisons it checks.
\end{itemize}

\section{Reproducibility}

{\raggedright
Scripts and JSON artifacts (public checkpoints, fixed seeds):
\texttt{census\_background\_reference.py} (\S\ref{sec:object},
\S\ref{sec:function}),
\texttt{census\_background\_identity.py} (\S\ref{sec:identity},
co-removal),
\texttt{census\_reference\_mechanism.py} (\S\ref{sec:bias},
\S\ref{sec:maintenance}),
\texttt{census\_checkpoint\_trajectory.py},
\texttt{census\_formation\_microscopy.py} and
\texttt{census\_formation\_prelude.py} (\S\ref{sec:formation},
Pythia-410M; the last is the sub-512 resolution check),
\texttt{induction\_gate\_coreg.py} (\S\ref{sec:formation}, the 160M
finer-grid duty-vs-induction time-course) and
\texttt{train\_noattn\_sparsification.py} (\S\ref{sec:formation}, the
160M model and its attention-disabled variant),
\texttt{census\_reference\_geometry.py} (geometric comparisons,
Related work),
\texttt{census\_reference\_geometry\_mahalanobis.py} (variance-aware
recomputation of those comparisons, with per-metric empirical nulls),
\texttt{census\_reference\_operating\_point.py} (\S\ref{sec:object},
the depth profile of the mean's share of the stream and the post-norm
reference),
\texttt{census\_reference\_statistic\_nulls.py} (\S\ref{sec:object},
\S\ref{sec:bias}: the permuted-offset and twin-injection nulls for the
duty correlation, and the cross-layer cosine and transfer matrices),
\texttt{census\_reference\_duty\_controls.py} (\S\ref{sec:bias}, the
random-direction, read-norm and bias-only controls),
\texttt{atlas\_backbone.py} (\S\ref{sec:function}, context-resolved
removal),
\texttt{census\_noise\_floor.py} and
\texttt{census\_noise\_zero\_mean.py} (\S\ref{sec:function}, the
matched-energy noise floor and the direction-matched injection),
\texttt{census\_crossmodel\_causal.py} (cross-model dose-response and
entropy response),
\texttt{census\_bias\_migration\_scale.py} (Table~\ref{tab:bias_xmodel}),
\texttt{census\_bias\_migration\_formation.py} (coupling time-course
over checkpoints),
\texttt{census\_bias\_migration\_nonlinearity.py} (eight-model scan,
OPT boundary),
\texttt{census\_reference\_norm\_decomp.py} (residual-versus-LayerNorm-bias
decomposition),

\texttt{census\_reference\_position\_shape.py} (\S\ref{sec:object}, the
prose, code and repeated-token reference cosines),
\texttt{census\_pass1\_scan.py} and \texttt{census\_drive\_ledger.py}
(the duty and reference-share figures of the introduction),
\texttt{census\_readweight\_factorization.py} (per-gate anatomy),
\texttt{census\_reference\_variance\_sources.py} (what holds the
fraction, \S\ref{sec:object}), \texttt{census\_reference\_mixture.py}
(token-class structure, \S\ref{sec:object}),
\texttt{census\_reference\_bias\_and\_depth.py} and
\texttt{census\_reference\_bias\_share\_xmodel.py} (the parameter share,
\S\ref{sec:bias}),
\texttt{census\_reference\_text\_statistics.py},
\texttt{census\_sigma\_dose\_response.py} and
\texttt{census\_duty\_at\_task\_position.py} (the input-side
demonstration and its scope, \S\ref{sec:function}),
\texttt{census\_reference\_horizon.py} (the sequence-position horizon).
Code and
data are available at
\url{https://github.com/EvidentSolutions/llm-interp/tree/main/background}
and archived at \url{https://doi.org/10.5281/zenodo.21498411}.
\par}

\section*{Use of AI Assistants}

Large language models were used as assistive tools for coding,
running experiments, and drafting text. All research questions,
experimental design, and reported claims were directed and verified by
the author, who takes full responsibility for the content.

\bibliographystyle{plainnat}
\bibliography{paper_background}

\end{document}